\documentstyle[aclap]{article}

\makeatletter

\newcounter{enums}

\newdimen\widelabel
\def\enumsentence{\@ifnextchar[{\@enumsentence}
{\refstepcounter{enums}\@enumsentence[(\theenums)]}}

\long\def\@enumsentence[#1]#2{\begin{list}{}{%
\advance\leftmargin by\widelabel \advance\labelwidth by \widelabel}
\item[#1] #2
\end{list}}

\newcounter{tempcnt}

\def\@item[#1]{\if@noparitem \@donoparitem
  \else \if@inlabel \indent \par \fi
         \ifhmode \unskip\unskip \par \fi 
         \if@newlist \if@nobreak \@nbitem \else
                        \addpenalty\@beginparpenalty
                        \addvspace\@topsep \addvspace{-\parskip}\fi
           \else \addpenalty\@itempenalty \addvspace\itemsep 
          \fi 
    \global\@inlabeltrue 
\fi
\everypar{\global\@minipagefalse\global\@newlistfalse 
          \if@inlabel\global\@inlabelfalse \hskip -\parindent \box\@labels
             \penalty\z@ \fi
          \everypar{}}\global\@nobreakfalse
\if@noitemarg \@noitemargfalse \if@nmbrlist \refstepcounter{\@listctr}\fi \fi
\setbox\@tempboxa\hbox{\makelabel{#1}}%
\global\setbox\@labels
 \hbox{\unhbox\@labels \hskip \itemindent
       \hskip -\labelwidth \hskip -\labelsep 
       \ifdim \wd\@tempboxa >\labelwidth 
                \box\@tempboxa
          \else \hbox to\labelwidth {\unhbox\@tempboxa}\fi
       \hskip \labelsep}\ignorespaces}

\newcounter{enumsi}

\newdimen\eeindent
\def\@mklab#1{\hfil#1}
\def\enummklab#1{\hfil(\eelabel)\hbox to \eeindent{\hfil#1}}
\def\enummakelabel#1{\enummklab{#1}\global\let\makelabel=\@mklab}
\def\toplabel#1{{\edef\@currentlabel{\p@enums\theenums}\label{#1}}}

\def\eenumsentence{\@ifnextchar[{\@eenumsentence}
{\refstepcounter{enums}\@eenumsentence[\theenums]}}

\long\def\@eenumsentence[#1]#2{\def\eelabel{#1}\let\holdlabel\makelabel%
\begin{list}{\alph{enumsi}.}{\usecounter{enumsi}%
\advance\leftmargin by \eeindent \advance\labelwidth by \eeindent%
\let\makelabel=\enummakelabel}
#2
\end{list}\let\makelabel\holdlabel}

\def\modsmalltree{\@ifnextchar[{\@modsmalltree}{\@modsmalltree[2]}}

\long\def\@modsmalltree[#1]#2#3{{\def\mc##1##2{%
\multicolumn{##1}{c}{\def\arraystretch{1}##2}}%
\def\arraystretch{#1}%
\def\ns{\def\arraystretch{1}}%
\setbox0=\hbox{\begin{tabular}[t]{@{}*{#2}{c}@{}}
#3
\end{tabular}}%
\dimen0=\ht0
\advance\dimen0 by -\arraystretch \ht\strutbox
\advance\dimen0 by \ht\strutbox
\ht0=\dimen0
\dimen0=\dp0
\advance\dimen0 by -\arraystretch \dp\strutbox
\advance\dimen0 by \dp\strutbox
\dp0=\dimen0
\leavevmode\box0}}

\def\clap#1{\hbox to 0pt{\hss#1\hss}}
\def\orlap#1#2{\vbox to0pt{\vss\hbox to 0pt{#2\hss}\hbox{\vrule height#1
width0pt depth0pt}}}
\def\ollap#1#2{\vbox to0pt{\vss\hbox to 0pt{\hss#2}\hbox{\vrule height#1
width0pt depth0pt}}}
\def\oclap#1#2{\vbox to0pt{\vss\hbox to 0pt{\hss#2\hss}\hbox{\vrule height#1
width0pt depth0pt}}}

\def\evnup{\@ifnextchar[{\@evnup}{\@evnup[0pt]}}

\def\@evnup[#1]#2{\setbox1=\hbox{#2}%
\dimen1=\ht1 \advance\dimen1 by -.5\baselineskip%
\advance\dimen1 by -#1%
\leavevmode\lower\dimen1\box1}

\makeatother

\let\+=\&

\renewcommand{\&}{\, \wedge \,}

\newcommand{\1}{_{1}}
\newcommand{\2}{_{2}}

\newcommand{\4}{_{4}}

\title{A Theory of Parallelism and the Case of VP Ellipsis}

\author{Jerry R. Hobbs and Andrew Kehler \\
Artificial Intelligence Center \\
SRI International \\ 
333 Ravenswood Avenue \\
Menlo Park, CA 94025 \\ 
{\tt \{hobbs,kehler\}@ai.sri.com}}

\begin{document}
\maketitle
\vspace{-0.5in}

\begin{abstract}
We provide a general account of parallelism in discourse, and
apply it to the special case of resolving possible readings for instances
of VP ellipsis.  We show how several problematic examples 
are accounted for in a natural and straightforward fashion.
The generality of the approach makes it directly
applicable to a variety of other types of ellipsis and reference.
\end{abstract}

\section{The Problem of VP Ellipsis}

VP ellipsis has received a great deal of attention in theoretical and
computational linguistics \cite[inter
alia]{Asher:93,Crouch:95,Dalrympl:91a,FiengoMay:94,Gawron:90a,Hardt:92a,Kehler:93a,Lappin:90a,Prust:92,Sag:76a,Webber:78a,Williams:77a}.
The area is a tangled thicket of examples in which readings are 
mysteriously missing and small changes reverse judgments.  It
is a prime example of a phenomenon at the boundary between syntax and
pragmatics.

VP ellipsis is exemplified in sentence~(\ref{simple-case}).

\enumsentence{\label{simple-case} John revised his paper before the
teacher did.}
This sentence has two readings, one in which the
teacher revised John's paper (the {\it strict} reading), and one in
which the teacher revised his own paper (the {\it sloppy} reading).
Obtaining an adequate account of strict/sloppy ambiguities has been a
major focus of VP ellipsis research.  This is 
challenging because not all examples are as simple as 
sentence~(\ref{simple-case}).  In fact, sentence~(\ref{simple-case})
is the first main clause of one of the more problematic cases in the
literature:

\enumsentence{\label{five-readings-case}
 John revised his paper before the teacher did, and Bill did too.}

Whereas one might expect there to be as many as six readings for this
sentence, Dalrymple~et~al.~\shortcite[henceforth~DSP]{Dalrympl:91a}
note that it has only five readings; the reading is absent in which

\enumsentence{John revised John's paper before the teacher revised
John's paper, and Bill revised John's paper before the teacher revised
Bill's paper. \label{five-sentence-readings-missing}}
\vspace*{-.15in}
Previous analyses have either generated too few or too many
readings, or have required an appeal to additional
processes or constraints external to the actual resolution process
itself.

Examples like (\ref{five-readings-case}) test the adequacy of an
analysis at a fine-grained level of detail.  Other examples test the
generality of an analysis, in terms of its ability to account for
phenomena similar to VP ellipsis and to interact with other
interpretation processes that may come into play.  For instance,
strict/sloppy ambiguities are not restricted to VP ellipsis, but are
common to a wide range of constructions that rely on parallelism
between two eventualities, some of which are listed in
Table~\ref{phenomena-table}.
{\small
\begin{table*}[t]
\begin{center}
\begin{tabular}{|l|l|} \hline 
Phenomenon & Example \\ \hline \hline
`Do It' Anaphora & John revised his paper before Bill did
it. \\ \hline
`Do So' Anaphora & John revised his paper and Bill did so too. \\ \hline

Stripping & John revised his paper, and Bill too. \\  \hline

Comparative Deletion & John revised his paper more quickly than
Bill. \\  \hline 

`Same As' Reference & John revised his paper, and Bill did
the same. \\ \cline{2-2}
& John revised his paper, and the teacher followed suit. \\ \hline

`Me Too' Phenomena & A: John revised his paper.
                      \\ 

& B: Me too./Ditto. \\ \hline

`One' Anaphora & John revised a paper of his, and Bill 
revised one too. \\ \hline

Lazy Pronouns & The student who revised his paper did better than \\  &
the student who handed it in as is. \\ \hline

Anaphoric Deaccenting & John said he called his teacher an idiot, \\ & 
and Bill {\small \sl said he insulted his teacher} too.  \\ \hline

Focus Phenomena & Only John revised his paper. \\ \hline
\end{tabular}
\caption{\label{phenomena-table}
Phenomena Giving Rise to Sloppy Interpretations}
\end{center}
\vspace*{-.15in}
\end{table*}
}
Given the ubiquity of strict/sloppy ambiguities, one would
expect these to be a by-product of general discourse
resolution mechanisms and not mechanisms specific to
VP ellipsis.  Any account applying only to the latter would 
miss an important generalization. 

In this paper, we give an account of resolution rooted in a general
computational theory of parallelism.  We demonstrate the depth of our
approach by showing that unlike previous approaches, the algorithm
generates the correct five readings for
example~(\ref{five-readings-case}) without appeal to additional
mechanisms or constraints. We also discuss how other `missing
readings' cases are accounted for.  We show the generality of the
approach by demonstrating its handling of several other examples that
prove problematic for past approaches, including a {\it
source-of-ellipsis paradox}, so-called {\it extended parallelism}
cases, and {\it sloppy readings with events} cases.  Of
the phenomena in Table~\ref{phenomena-table}, we briefly
discuss the algorithm's handling of {\it lazy pronoun} cases.

\section{A Theory of Parallelism}
\label{parallelism-theory-section}

\paragraph{The Theory}
A clause conveys a property or eventuality, or describes a situation,
or expresses a proposition.  We use the term ``property'' to
cover all of these cases.  A property consists of a predicate applied
to a number of arguments.  We make use of a duality between properties
having a number of arguments, and arguments having a number of
properties.  Parallelism is characterized in terms of a co-recursion
in which the similarity of properties is defined in terms of the
similarity of arguments, and the similarity of arguments is defined in
terms of the similarity of properties.\footnote{This account is a
elaboration of treatments of parallelism by
Hobbs~\shortcite{Hobbs:79,Hobbs:85b} and
Kehler~\shortcite{Kehler:Thesis}.}

Two fragments of discourse stand in a parallel relation if they
describe similar properties.  Two properties are similar if two
corresponding properties can be inferred from them in which the
predicates are the same and the corresponding pairs of arguments are
either coreferential or similar.

\vspace*{.05in}

\noindent $Similar[p_1'(e_1,x_1,\ldots,z_1), p_2'(e_2,x_2,\ldots,z_2)]$: \\
\hspace*{.5cm} $p_1'(e_1,x_1,\ldots,z_1) \supset p'(e_1,x_1,\ldots,z_1)$ and \\
\hspace*{.5cm} $p_2'(e_2,x_2,\ldots,z_2) \supset p'(e_2,x_2,\ldots,z_2)$, where \\
\hspace*{1cm} $\mbox{\it Coref}\/(x_1,\ldots,x_2,\ldots)$ or $Similar[x_1,x_2]$, \\
\hspace*{1cm} $\ldots$ \\
\hspace*{1cm} $\mbox{\it Coref}\/(z_1,\ldots,z_2,\ldots)$ or $Similar[z_1,z_2]$
\vspace*{.05in}

\noindent Two arguments are similar if their other, ``inferentially
independent'' properties are similar.
\vspace*{.05in}

\noindent $Similar[x_1,x_2]$: \\
\hspace*{.5cm} $Similar[p_1'(\ldots,x_1,\ldots), p_2'(\ldots,x_2,\ldots)]$, \\ 
\hspace*{.5cm} $\ldots$ \\
\hspace*{.5cm} $Similar[q_1'(\ldots,x_1,\ldots), q_2'(\ldots,x_2,\ldots)]$
%

\noindent The constructed mapping between pairs of arguments must be preserved
and remain one-to-one.

There are three ways the recursion can bottom out.  We can run out of
new arguments in properties.  We can run out of new, inferentially
independent properties of arguments.  And we can ``bail out'' of
proving similarity by proving or assuming coreference between the two
entities.

Two properties are {\it inferentially independent} if neither can be
derived from the other.  Given a knowledge base $K$ representing the
mutual knowledge of the participants in the discourse, properties
$P\1$ and $P\2$ are inferentially independent if neither $K, P\1
\vdash P\2$ nor $K, P\2 \vdash P\1$.  This rules out the case in
which, for example, the fact that John and Bill are both persons would
be used to establish their similarity when the fact that they are both
men has already been used.  Inferential independence is generally
undecidable, but in practice this is not a problem.  In discourse
interpretation, all we usually know about an entity is the
small set of properties presented explicitly in the text itself.  We
may take these to be inferentially independent and look for no
further properties, once properties inferrable from these have
been used in establishing the parallelism.

Similarity is a matter of degree.  The more corresponding pairs of
inferentially independent properties that are found, and the more
contextually salient those properties are, the stronger the
similarity.  In a system which assigns different costs to proofs (e.g.,
Hobbs et al.~\shortcite{HobbsEtAl:93a}), the more costly the proofs
required to establish similarity are, the less similar the properties
or arguments should seem.  Interpretations should seek to maximize
similarity.

This account of parallelism is semantic in the sense that it depends
on the content of the discourse rather than directly on its form.  But
syntax plays an implicit role.  When seeking to establish the
parallelism between two clauses, we must begin with the ``top-level''
properties; this is generally determined by the syntactic structure of
the clause.  Then the co-recursion through the arguments and
properties normally mirrors the syntactic structure of the sentence.
However, features of syntax that are not manifested in logical form
are not taken into account.  

\paragraph{An Example} 
To illustrate that the theory has applicability well beyond the
problem of VP ellipsis, we present an example of semantic parallelism
in discourse.  It comes from an elementary physics textbook, and is
worked out in essentially the same manner in Hobbs~\shortcite{Hobbs:79}.

\enumsentence{A ladder weighs 100 lb with its center of gravity 10 ft from
	the foot, and a 150 lb man is 10 ft from the top.}
We will assume ``the foot'' has been identified as the foot of the
ladder.  Because it is a physics problem, we must reduce the two clauses to
statements about forces acting on objects with magnitudes in a
direction at a point in the object:

\begin{verse}
$ force(w\1,L,d\1,x\1); force(w\2,y,d\2,x\2) $ 
\end{verse}

In the second clause we do not know that the man is standing
on the ladder---he could be on the roof---and we do not know what ``the
top'' is the top of.  These facts fall out of recognizing the
parallelism.  

The procedure for establishing parallelism is illustrated in
Figure~\ref{force-figure}, in which 
parallel elements are placed on the same line.
\begin{figure*}[ht]
\begin{center}
\begin{tabular}{ll}
$force(w_1,L,d_1,x_1)$ & \hspace{1in}   $force(w_2,y,d_2,x_2)$ \\
\hspace{.2in}   $w_1: lb(w_1,100)$  & \hspace{1.2in}  $w_2: lb(w_2,150)$ \\
\hspace{.4in}   $L:  ladder(L)$   & \hspace{1.4in} $y: \Rightarrow Coref(y,...,L,...)$ \\
\hspace{.4in}   $d_1: Down(d_1)$    & \hspace{1.4in}   $d_2: Down(d_2)$ \\
\hspace{.4in}   $x_1: distance(x_1,f,20ft)$  & \hspace{1.4in} $x_2: distance(x_2,t,10ft)$ \\
\hspace{.6in}   $f: foot(f,L) \Rightarrow end(f,L)$  & \hspace{1.6in}  $t: top(t,z) \Rightarrow end(t,z)$ \\
\hspace{.8in}   $L:$  &  \hspace{1.8in} $z: \Rightarrow Coref(z,...,L,...)$ \\
\end{tabular}
\end{center}
\caption{\label{force-figure} Example of Parallelism Establishment}
\vspace*{-.15in}
\end{figure*}
The {\it force} predicates are the same so there is no need to infer
further properties.  The first pair of arguments, $w\1$ and $w\2$ are
similar in that both are weights.  To make the second pair
of arguments similar, we can assume they are coreferential; as a
by-product, this tells us that the object the man's weight is acting
on is the ladder, and hence that the man is on the ladder.  The third
pair of arguments are both
downward directions.  The final pair of arguments, $x\1$
and $x\2$, are similar if their properties $distance(x\1,f,20\mbox{\it
ft})$
and $distance(x\2,t,10\mbox{\it ft})$ are similar.  These will be similar if
their previously unmatched pair of arguments $f$ and $t$ are similar.
This holds if their properties
$\mbox{\it foot}(f,L)$ and $top(t,z)$ are similar.  We infer
$end(f,L)$ and $end(t,z)$, since feet and tops are ends.  Finally, we
have to show $L$ and $z$ are similar.  We can do this by assuming 
they are coreferential.  This, as a by-product, tells us that the top
is the top of the ladder.

The use of inferences, such as ``a foot is an end'', means that this
theory is parametric on a knowledge base.  Different
sets of beliefs can yield different bases for parallelism and indeed
different judgments about whether parallelism occurs at all.

A crucial piece of our treatment of VP-ellipsis is the explicit
representation of coreference relations, denoted with the predicate
{\it Coref}.  We could use equalities
such as $y=L$, or since equals can be replaced by equals, simply
replace $y$ with $L$.  However, doing this would lose the distinction
between $y$ and $L$ under their corresponding descriptions.

Consequently, we introduce the relation \linebreak $\mbox{\it
Coref}\/(y,e_2,x,e_1)$ to express this coreferentiality.  This
relation says
that $y$ under the description associated with $e_2$ is coreferential
with $x$ under the description associated with $e_1$.  From this we
can infer $y = x$ but not $e_2 = e_1$, and the coreferentiality cannot
be washed out in substitution.  A constraint on the arguments of
{\it Coref} is that $e\1$ and $e\2$ be properties of $x$ and $y$
respectively.

The phenomenon of parallelism pervades discourse.  In addition to
straightforward examples of parallelism like the above, there are also
contrasts, exemplifications, and generalizations, which are defined in
a similar manner.  The interpretation of a number of syntactic
constructions depends on recognizing parallelism, including those cited
in Table~\ref{phenomena-table}.  In brief, our theory of parallelism
is not something we have introduced merely for the purpose of handling
VP ellipsis; it is needed for a wide range of sentential and
discourse phenomena.

\paragraph{Other Approaches Based on Parallelism}
Our aim in this paper is to present the theory of parallelism at an
abstract enough level that it can be embedded in any sufficiently
powerful framework.  By ``sufficiently powerful'' we mean that there
must be a formalization of the notion of inference, strength of
inference, and inferential independence, and there must be a
reasonable knowledge base.  In Hobbs and
Kehler~\shortcite{HobbsKehler:forthcoming}, we show how our approach
can be realized within the ``Interpretation as Abduction''
framework \cite{HobbsEtAl:93a}.  

There are at least two other treatments in which VP ellipsis is
resolved through a more general system of determining discourse parallelism,
namely, those of Pr\"{u}st \shortcite{Prust:92} and
Asher \shortcite{Asher:93}.

Pr\"{u}st \shortcite{Prust:92} gives an account of parallelism
developed within the context of the Linguistic Discourse Model theory
\cite{Scha:88a}.  Parallelism is
computed by determining the ``Most Specific Common Denominator'' of a
set of representations, which results from unifying the unifiable
aspects of those representations and generalizing over the others.  VP
ellipsis is resolved as a side effect of this unification.  The
representations assumed, called {\it syntactic/semantic structures},
incorporate both syntactic and semantic information about an
utterance.  One weakness of this approach is that it appears 
overly restrictive in the syntactic similarity that it requires.

Asher \shortcite{Asher:93} also provides an analysis of VP ellipsis in the
context of a theory of discourse structure and coherence, 
using an extension of Discourse Representation Theory.  The resolution
of VP ellipsis is driven by a need to maximize parallelism (or in some
cases, contrast) that is very much in the spirit of what we present.

Detailed comparisons with our approach are given with the examples
below.  In general, however, in neither of these approaches has enough
attention been paid to other interacting phenomena to explain the
facts at the level of detail that we do.

\section{VP Ellipsis:  A Simple Case}
\label{simple-case-section}

\begin{figure*}[ht]
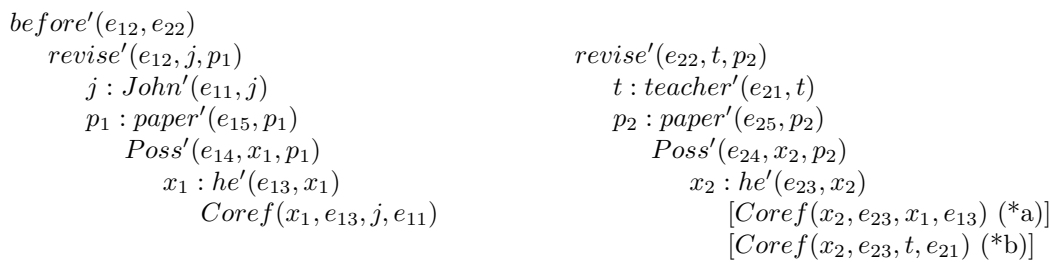

\begin{tabular}{ll}
\hspace{.3in} $before'(e_{12},e_{22})$ & \\
\hspace{.5in} $revise'(e_{12},j,p_1)$  & \hspace{.5in} $revise'(e_{22},t,p_2)$ \\
\hspace{.7in} $j: John'(e_{11},j)$ & \hspace{.7in} $t: teacher'(e_{21},t)$ \\
\hspace{.7in} $p_1: paper'(e_{15},p_1)$ & \hspace{.7in}  $p_2: paper'(e_{25},p_2)$ \\
\hspace{.9in} $Poss'(e_{14},x_1,p_1)$ & \hspace{.9in} $Poss'(e_{24},x_2,p_2)$ \\ 
\hspace{1.1in} $x_1: he'(e_{13},x_1)$ & \hspace{1.1in} $x_2: he'(e_{23},x_2)$ \\
\hspace{1.3in}  $Coref(x_1,e_{13},j,e_{11})$ & \hspace{1.3in}
[$Coref(x_2,e_{23},x_1,e_{13})$  (*a)] \\  
 & \hspace{1.3in}    [$Coref(x_2,e_{23},t,e_{21})$    (*b)] \\
\end{tabular}
\caption{\label{simple-target-figure}
Representations for Simple Case}
\vspace*{-.15in}
\end{figure*}

We first illustrate our approach on the simple case of VP ellipsis in
sentence~(\ref{simple-case}).  The representation for the antecedent clause
in our ``logical form''\footnote{The normally controversial term
``logical form'' is used loosely here, simply to capture the
information that the hearer must bear in mind, at least implicitly, in
interpreting texts such as sentence (\ref{simple-case}).} appears on
the left-hand side of Figure~\ref{simple-target-figure}.
Note that a {\it Coref} relation links $x_1$, the variable 
corresponding to ``he'' (eventuality $e_{13}$), to its antecedent $j$, the
entity described by ``John'' (eventuality $e_{11}$).

From the second clause we know there is an elided eventuality $e_{22}$
of unknown type $P$, the logical subject of which is the teacher $t$.

\begin{quote}
 $P(e_{22},t)$ \\ 
\hspace*{.2in} $t: teacher'(e_{21},t)$
\end{quote}
Because of the ellipsis, $e_{22}$ must stand in a parallel relation to
some previous eventuality; here the only candidate is John's revising
his paper ($e_{12}$).  To establish $\mbox{\it
Similar}\/(e_{12},e_{22})$,\footnote{\label{event-link-footnote} We
cannot establish coreference between the events because their agents
are distinct.  In other cases, however, the process can bail out
immediately in event coreference; consider the sentence ``John revised
his paper, smoking incessantly as he did.'' A {\it Coref} link is
established between the elided and antecedent events in the same way
as for pronouns. This symmetry accounts for another problematic case,
discussed in Section~\ref{other-examples-section}.} we need to show
that their corresponding arguments are similar.  John $j$ and the
teacher $t$ are similar by virtue of being persons.  The corresponding
objects $p_1$ and $p_2$ are similar if we take $p_2$ to be a paper and
to have a $Poss$ property similar to that of $p_1$.  The latter is
true if corresponding to the possessor $x_1$, there is an $x_2$ that
is similar to $x_1$.

In constructing the similarity between $x_2$ and $x_1$, we can either
take them to be coreferential (case *a) or prove them to be
similar by having similar properties, including having similar
dependencies established by {\it Coref} (case *b).  In the
former case, $x_2$ is coreferential with $x_1$ which is coreferential
with John $j$, giving us the strict reading.  In the latter case, we
must preserve the previously-constructed mapping between John $j$ (on
which $x_1$ is dependent) and the teacher $t$; thus $x_2$ is similar
to $x_1$ if taken to be coreferential with $t$, giving us the sloppy
reading.\footnote{It is also possible to ``bail out'' in coreference
between the papers $p_1$ and $p_2$; here we would get the strict
reading again.  However, consider if the example had said ``a paper of
his'' rather than ``his paper''.  The resulting sentence has two
strict readings, one in which both revised the same paper of John's
(generated by assuming coreference between the papers), and one in
which each revised a (possibly) different paper of John's (generated
by assuming coreference between the pronouns).}

\section{A Missing Readings Paradox}

Sentence~(\ref{simple-case}) is the antecedent clause for example
(\ref{five-readings-case}), one of the more problematic examples in
the literature.  Theoretically, this example could have as many as six
readings, paraphrased as follows:

\enumsentence{John revised John's paper before the teacher revised
John's paper, and Bill revised John's/Bill's paper before the teacher
revised John's/Bill's paper. }

\enumsentence{John revised John's paper before the teacher revised
the teacher's paper, and Bill revised John's/Bill's paper before the teacher
revised the teacher's paper. }
We follow DSP in claiming that this example has five readings, in
which the JJJB reading shown in (\ref{five-sentence-readings-missing})
is missing.\footnote{Each reading for this example contains four
descriptions of papers that were revised.  We use the notation JJJB to
represent the reading in which the first three papers are John's and
fourth is Bill's, corresponding to reading
(\ref{five-sentence-readings-missing}). Other uses of such notation
should be understood analogously.} DSP, who use this case as a
benchmark for theories of VP ellipsis, note that the methods of
Sag~\shortcite{Sag:76a} and Williams~\shortcite{Williams:77a} can be
seen to derive two readings, namely JJJJ and JTBT.  An analysis
proposed by Gawron~and~Peters~\shortcite{Gawron:90a}, who first
introduced this example, generates three readings (adding JJBB to the
above two), as does the analysis of Fiengo and
May~\shortcite{FiengoMay:94}.  A method that Gawron and Peters
attribute to Hans Kamp generates either four readings, including the
above three and JTJT, or all six readings.  DSP's analysis strictly
speaking generates all six readings; however, they appeal to
anaphor/antecedent linking relationships to eliminate the JJJB
reading.  However, these linking relationships are not a by-product of
the resolution process itself, but must be generated
separately.  Our approach derives exactly the correct five
readings.\footnote{The approach presented in
Kehler~\shortcite{Kehler:93a} also derives the correct five readings,
however, our method has advantages in its being more general and
better motivated.}

The antecedent clause is represented in
Figure~\ref{simple-target-figure}, and the expansion of the final VP
ellipsis is shown in Figure~\ref{five-readings-figure}.  In proving
similarity, each pronoun can be taken to be coreferential with its
parallel element (cases *a, *c and *e), or proven similar to it (cases
*b, *d, *f and *g).  If choice *a is taken in the second clause, then
the ``similarity'' choice in the fourth clause must be *f; if *b, then
*g.
\begin{figure*}[ht]
\begin{tabular}{ll}
\hspace{.1in} $before(e_{32},e_{42})$ \\
\hspace{.3in}   $revise'(e_{32},b,p_3)$  &   \hspace{.3in} $revise'(e_{42},t,p_4)$ \\
\hspace{.5in}  $b: Bill'(e_{31},b)$  &  \hspace{.5in}  $t: teacher'(e_{41},t)$ \\
\hspace{.5in}  $p_3: paper'(e_{35},p_3)$ & \hspace{.5in} $p_4: paper'(e_{45},p_4$)\\
\hspace{.7in}  $Poss'(e_{34},x_3,p_3)$    &   \hspace{.7in}   $Poss'(e_{44},x_4,p_4)$\\
\hspace{.9in} $x_3: he'(e_{33},x_3)$   &  \hspace{.9in}  $x_4: he'(e_{43},x_4)$\\
\hspace{1.1in} [(*c)             $Coref(x_3,e_{33},x_1,e_{13})$] & \hspace{1.1in}
[$Coref(x_4,e_{43},x_2,e_{23})$  (*e)]      \\
\hspace{1.1in} [(*d)             $Coref(x_3,e_{33},b,e_{31})$]   &  \hspace{1.1in}
[$Coref(x_4,e_{43},x_3,e_{33})$  (*f)] \\
   &  \hspace{1.1in} [$Coref(x_4,e_{43},t,e_{41})$  (*g)] \\
\end{tabular}
\caption{\label{five-readings-figure}
Representations for Five Readings Case}
\vspace*{-.15in}
\end{figure*}
If *a and *c are chosen, the JJJJ reading results.  If *a, *d, and *e
are chosen, the JJBJ reading results.  If *a, *d, and *f are chosen,
the JJBB reading results.  If *b and *c are chosen, the JTJT reading
results.  If *b and *d are chosen, the JTBT reading results.  Thus
taking all possible choices gives us all acceptable readings.

Now consider what it would take to obtain the *JJJB reading.  The
variable $x_3$ would have to be coreferential with John and $x_4$ with
Bill.  The former requirement forces us to pick case *c.  But then
case *e makes $x_4$ coreferential with either John or the teacher
(depending on how the first ellipsis was resolved).  Case *f makes
$x\4$ coreferential with John, and case *g makes it coreferential with
the teacher.  There is no way to get $x_4$ coreferential with Bill
once we have set $x_3$ to something other than Bill.

Neither Pr\"{u}st \shortcite{Prust:92} nor Asher \shortcite{Asher:93}
discuss this example.  In extrapolating from the analyses Pr\"{u}st
gives, we find that his analysis generates only two of the five
readings.  Briefly, if the first ellipsis is resolved to the strict
reading, then the JJJJ reading is possible.  If the first ellipsis is
resolved to the sloppy reading, then only the JTBT reading is
possible.  Asher's account, extrapolating from an example he
discusses (p. 371),  may generate as many as six readings,
including the missing reading.  This reading results from the manner
in which the strict reading for the first ellipsis is generated---the
final clause pronoun is resolved with the entity specified by the
subject of the antecedent clause, whereas our algorithm creates a
dependency between the pronoun and its parallel element in the
antecedent clause.  Our mechanism is more natural because of the
alignment of parallel elements between clauses when establishing
parallelism, and it is this property which results in the
underivability of the missing reading.

\section{A Source-of-Ellipsis Paradox}

DSP identify two kinds of analysis in the VP ellipsis literature.  In
{\it identity-of-relations} analyses
\cite[inter~alia]{Sag:76a,Williams:77a,Gawron:90a,FiengoMay:94}
strict/sloppy readings arise from an ambiguity in the antecedent VP
derivation.  The ambiguity in the ellipsis results from copying each
possibility.  In {\it non-identity} approaches
\cite[inter~alia]{Dalrympl:91a,Kehler:93a,Crouch:95} strict/sloppy
readings result from a choice point within the resolution algorithm.
Our approach falls into this class.

Non-identity approaches are supported by examples such as
(\ref{cascaded-ellipsis}), which has reading~(\ref{cascaded-reading}).

\enumsentence{John realizes that he is a fool, but Bill does not, even
though his wife does.\label{cascaded-ellipsis} \cite{Dahl:72a}}

\enumsentence{\label{cascaded-reading}
John realizes that John is a fool, but Bill does not
realize that Bill is a fool, even though Bill's wife realizes
Bill is a fool.} 
Example (\ref{cascaded-ellipsis}) contains two ellipses.
Reading~(\ref{cascaded-reading}) results from the second clause
receiving a sloppy interpretation from the first, and the third clause
receiving a strict interpretation from the second.  An
identity-of-relations analysis, however, predicts that
this reading does not exist.  Because the second clause will only have
the sloppy derivation received from the first, the strict derivation
that the third clause requires from the second will not be present.

However, in defending their identity-of-relations approach,
Gawron~and~Peters~\shortcite{Gawron:90a} note that a non-identity
account predicts that sentence (\ref{gawron-sentence}) has the
(nonexistent) reading given in (\ref{gawron-sentence-non-reading}).

\enumsentence{John revised his paper before Bill did, but 
after the teacher did.\label{gawron-sentence}}

\enumsentence{John revised John's paper before Bill revised Bill's paper, but 
after the teacher revised John's paper.\label{gawron-sentence-non-reading}}
In this case, the first clause is the antecedent for both ellipses.
These two examples create a paradox; apparently neither type
of analysis (nor any previous analyses we are aware of) can explain
both.

Our analysis accounts for both examples through a
mutually-constraining interaction of parallelisms.  Example
(\ref{cascaded-ellipsis}) is fairly straightforward, so we focus on
example~(\ref{gawron-sentence}).  Let us refer to the clauses as
clauses 1, 2, and 3.  Because clauses 2 and 3 are VP-elliptical, we
must establish a parallelism between each of them and clause 1.  In
addition, the contrast relation signalled by ``but'' is justified by
the contrasting predicates ``before'' and ``after'', provided their
corresponding pairs of arguments are similar.  Their first arguments
are similar since they are identical---clause 1.  Then we also have to
establish the similarity of their second arguments---clause 2 and
clause 3.  Thus, three mutually constraining parallelisms must be
established: 1 -- 2, 1 -- 3, and 2 -- 3.
\begin{figure*}[ht]
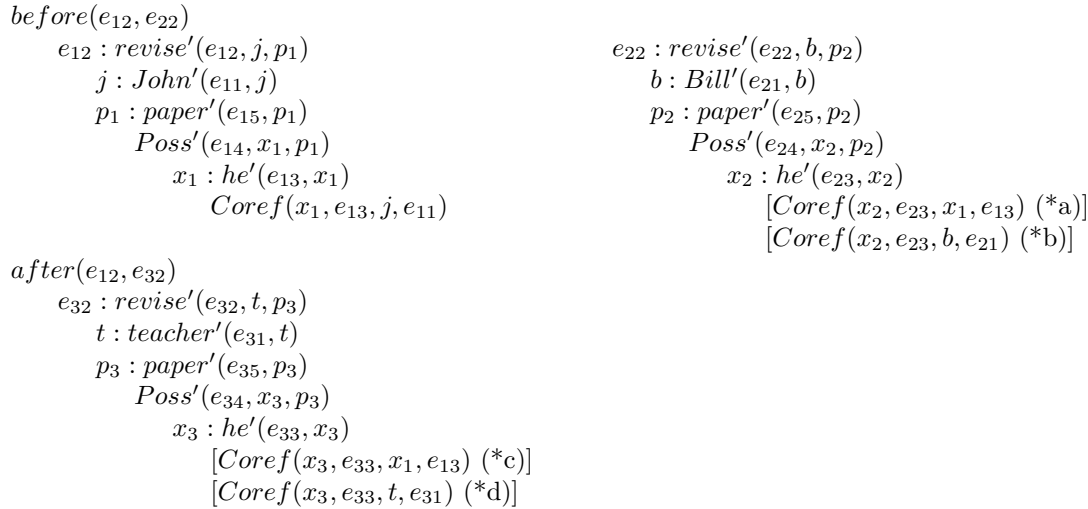

\begin{tabular}{ll}

$before(e_{12},e_{22})$ \\
\hspace{.2in} $e_{12}: revise'(e_{12},j,p_1)$ & \hspace{.2in} $e_{22}: revise'(e_{22},b,p_2)$ \\
\hspace{.4in}    $j: John'(e_{11},j)$  & \hspace{.4in} $b: Bill'(e_{21},b)$ \\
\hspace{.4in}    $p_1: paper'(e_{15},p_1)$ & \hspace{.4in} $p_2: paper'(e_{25},p_2)$ \\
\hspace{.6in}    $Poss'(e_{14},x_1,p_1)$  & \hspace{.6in} $Poss'(e_{24},x_2,p_2)$ \\
\hspace{.8in}    $x_1: he'(e_{13},x_1)$  & \hspace{.8in}  $x_2: he'(e_{23},x_2)$ \\
\hspace{1in}   $Coref(x_1,e_{13},j,e_{11})$ & \hspace{1in}
[$Coref(x_2,e_{23},x_1,e_{13})$  (*a)] \\
       & \hspace{1in}  [$Coref(x_2,e_{23},b,e_{21})$   (*b)] \\ 
$after(e_{12},e_{32})$ \\
\hspace{.2in} $e_{32}: revise'(e_{32},t,p_3)$ & \\
\hspace{.4in}       $t: teacher'(e_{31},t)$ & \\
\hspace{.4in}       $p_3: paper'(e_{35},p_3)$ & \\
\hspace{.6in}           $Poss'(e_{34},x_3,p_3)$ & \\
\hspace{.8in}              $x_3: he'(e_{33},x_3)$ & \\
\hspace{1in}              [$Coref(x_3,e_{33},x_1,e_{13})$  (*c)] & \\
\hspace{1in}              [$Coref(x_3,e_{33},t,e_{31})$  (*d)] & \\
\end{tabular}
\caption{\label{source-paradox-figure}
Representations for the Source-of-Ellipsis Paradox}
\vspace*{-.15in}
\end{figure*}

In Figure~\ref{source-paradox-figure}, cases *a and *b arise from the
coreference and similarity options when establishing the parallelism
between clauses 1 and 2, and cases *c and *d from the parallelism
between clauses 1 and 3.  However, because parallelism is also
required between clauses 2 and 3, we cannot choose these options
freely.  If we choose case *a, then we must choose case *c, giving us
the JJJ reading.  If we choose case *b, then we must choose case *d,
giving us the JBT reading.  Because of the mutual constraints of the
three parallelisms, no other readings are possible.  This is exactly
the right result.

Pr\"{u}st \shortcite{Prust:92} essentially follows Sag's
\shortcite{Sag:76a} treatment of strict and sloppy readings, which,
like other identity-of-relations analyses, will not generate the
reading of the cascaded ellipsis sentence (\ref{cascaded-ellipsis})
shown in (\ref{cascaded-reading}).  While the approach will correctly
predict the lack of reading (\ref{gawron-sentence-non-reading}) for
sentence (\ref{gawron-sentence}), it does so for the wrong reason.
Whereas ellipsis resolution does not permit such readings in any
circumstance in his account, we claim that the lack of such readings
for sentence (\ref{gawron-sentence}) is due to constraints imposed by
multiple parallelisms, and not because of the correctness of
identity-of-relations analyses.

Asher's \shortcite{Asher:93} analysis falls into the non-identity
class of analyses, and therefore makes the correct predictions for
sentence (\ref{cascaded-ellipsis}).  While he does not discuss the
contrast between this case and sentence (\ref{gawron-sentence}), we do
not see any reason why his framework could not accommodate our
solution.

\section{Other Examples}
\label{other-examples-section}

\paragraph{Missing Readings with Multiple Pronouns}  

Dahl~\shortcite{Dahl:74a} noticed that sentence (\ref{3-of-4}) has
only three readings instead of the four one might expect. The
reading {\it Bill said that John revised Bill's paper} is missing.

\enumsentence{John said that he revised his paper, and Bill did
too.\label{3-of-4}} 

In contrast, the similar sentence given in (\ref{all-4-readings})
appears to have all four readings.

\enumsentence{John said that his teacher revised his paper, and Bill did too.
\label{all-4-readings}}

The readings derived by our analysis depend on the {\it Coref}
relations that hold between the coreferring noun phrases in the antecedent
clauses.  For sentence (\ref{3-of-4}), the correct readings result if
{\it his} is linked to {\it he} and {\it he} to {\it John}; for
sentence (\ref{all-4-readings}), the correct readings result if both
pronouns are linked to {\it John}.  Other cases in the literature
indicate that the situation is more complicated than might initially
be evident.  Handling these cases requires an account
of how such dependencies are established, which we discuss in Hobbs
and Kehler~\shortcite{HobbsKehler:forthcoming}.

\paragraph{Extended Parallelism}

In some cases, the elements involved in a sloppy reading may not
be contained in the minimal clause containing the ellipsis.

\enumsentence{\label{extended-case} John told a man that Mary likes
him, and Bill told a boy that Susan does.\footnote{This example is due
to Pr\"{u}st \shortcite{Prust:92}, whose approach successfully handles
this example.}}

Although the antecedent clause for ``Susan does'' is ``Mary likes
him'', there is a sloppy reading in which ``Bill told a boy that Susan
likes Bill''.  This fact is problematic for accounts of VP
ellipsis that operate only within the minimal clauses.  These readings
are predicted by our account, as John and Bill are 
parallel in the main clauses.

\paragraph{Lazy Pronouns}

``Lazy pronouns'' can be accounted for similarly.  In 

\enumsentence{The man who gives his paycheck to his wife is wiser than
		the man who gives it to his mistress.
		\cite{Karttunen:69} \label{lazy-pronoun-ex}}
the pronoun {\it it} does not refer to the first man's paycheck but the 
second's.  

In text, {\it it} normally requires an explicit, coreferring antecedent. 
However, the parallelism between the clauses licenses a sloppy reading
via the similarity option.  The real world fact that to give something
to someone, you first must have it, leads to a strong preference for
the sloppy reading.

It is necessary to have parallelism in order to license the lazy pronoun
reading.  If we eliminate the possibility of parallelism, as in 

\enumsentence{
	John revised his paper, and then Bill handed it in.
}
the lazy pronoun reading is not available, even though the
have-before-give constraint is not satisfied.  To interpret this
sentence, we are more likely to assume an unmentioned transfer event
between the two explicit events.

\paragraph{Sloppy Readings with Events}

Sentence (\ref{sloppy-event}) has a ``sloppy'' reading in which the
second main clause means ``I will kiss you even if you don't want me to kiss
you.''  

\enumsentence{\label{sloppy-event} I will help you if you want me to,
but I will kiss you even if you don't.\footnote{Mark Gawron, p.c.,
attributed to Carl Pollard.}}

Deriving this reading requires a {\it Coref} relation between the
elided event and its antecedent in the first main clause, which is obtained
when our algorithm bails out in event coreference (see footnote
\ref{event-link-footnote}).  Then in expanding the VP ellipsis in the
second main clause, taking the similarity option for the event 
generates the desired reading.

\paragraph{Inferentially-Determined Antecedents}
\label{Webberex}

Webber~\shortcite{Webber:78a} provides several examples in which the
antecedent of an ellipsis is derived inferentially:

\enumsentence{Mary wants to go to Spain and Fred wants to go to Peru, but
because of limited resources, only one of them can.\label{webber1}} 

Our account of parallelism applies twice in handling this example,
once in creating a complex antecedent from recognizing the
parallelism between the first two clauses, and again in resolving the
ellipsis against this antecedent. 
Hobbs and Kehler~\shortcite{HobbsKehler:forthcoming}
describe the analysis of this case as well as others
involving quantification.

\section{Summary}

We have given a general account of parallelism in discourse and
applied it to the special case of resolving possible readings for
instances of VP ellipsis.  In doing so, we showed how a variety of
examples that have been problematic for previous approaches are
accounted for in a natural and straightforward fashion.
Furthermore, the generality of the approach
makes it directly applicable to a variety of other types of ellipsis
and reference in natural language.

\section*{Acknowledgements}

The authors thank Mark Gawron, David Israel, and three anonymous
reviewers for helpful comments.  This research was supported by
National Science Foundation/Advanced Research Projects Agency Grant
IRI-9314961.

\end{document}